\documentclass{article}
\usepackage{arxiv}

\usepackage[utf8]{inputenc} %
\usepackage[T1]{fontenc}    %
\usepackage{hyperref}       %
\usepackage{url}            %
\usepackage{booktabs}       %
\usepackage{amsfonts}       %
\usepackage{nicefrac}       %
\usepackage{microtype}      %
\usepackage{graphicx}
\usepackage[numbers]{natbib}
\usepackage{doi}
\usepackage{pifont}

\usepackage{xcolor}

\definecolor{navyblue}{HTML}{0071BC}
\definecolor{hotpink}{HTML}{FF0080}

\definecolor{oai-white}{HTML}{FFFFFF}
\definecolor{oai-black}{HTML}{000000}
\definecolor{oai-red}{HTML}{FF4500}
\definecolor{oai-green}{HTML}{51DA4C}
\definecolor{oai-blue}{HTML}{0000FF}
\definecolor{oai-yellow}{HTML}{FFF639}
\definecolor{oai-magenta}{HTML}{FF45FF}
\definecolor{oai-cyan}{HTML}{00FFFF}
\definecolor{oai-orange}{HTML}{FE7600}
\definecolor{oai-violet}{HTML}{8A2BE2}
\definecolor{oai-brown}{HTML}{A0522D}
\definecolor{oai-green-050}{HTML}{F4FFF4}
\definecolor{oai-green-100}{HTML}{E9FFE8}
\definecolor{oai-green-200}{HTML}{D9FFD8}
\definecolor{oai-green-300}{HTML}{C9FFC7}
\definecolor{oai-green-400}{HTML}{A6FFA3}
\definecolor{oai-green-500}{HTML}{7CF178}
\definecolor{oai-green-600}{HTML}{51DA4C}
\definecolor{oai-green-700}{HTML}{3FA93B}
\definecolor{oai-green-800}{HTML}{2D712A}
\definecolor{oai-green-900}{HTML}{193718}
\definecolor{oai-gray-000}{HTML}{FFFFFF}
\definecolor{oai-gray-100}{HTML}{FAFAFA}
\definecolor{oai-gray-200}{HTML}{F5F5F5}
\definecolor{oai-gray-300}{HTML}{E5E5E5}
\definecolor{oai-gray-400}{HTML}{FFB7A4}
\definecolor{oai-gray-500}{HTML}{CDCDCD}
\definecolor{oai-gray-600}{HTML}{A8A8A8}
\definecolor{oai-gray-700}{HTML}{747474}
\definecolor{oai-gray-800}{HTML}{393939}
\definecolor{oai-gray-900}{HTML}{000000}
\definecolor{visual}{HTML}{A50E0E}       
\definecolor{linguistic}{HTML}{174EA6}   
\definecolor{relational}{HTML}{E37400}   
\definecolor{egocentric}{HTML}{0D652D}

\definecolor{gain}{HTML}{009900}
\colorlet{gain_bg}{oai-green-050}
\colorlet{linecolor1}{oai-gray-200}
\colorlet{linecolor2}{oai-gray-300}
\colorlet{drop}{oai-red}

\usepackage{colortbl} %
\usepackage{multirow}
\usepackage{xspace} %
\usepackage{amsmath}
\usepackage{cleveref}       %
\usepackage{soul}
\usepackage{authblk}

\crefname{section}{Sec.}{Secs.}

\crefname{figure}{Fig.}{Figs.}

\crefname{table}{Tab.}{Tabs.}

\crefname{equation}{Eq.}{Eqs.}

\makeatletter
\renewcommand\AB@affilsepx{\protect\Affilfont\hspace{1em}}
\makeatother

\newcommand{\ours}{Q-GeoMem\xspace}

\newcommand{\localmem}{FGCB\xspace}
\newcommand{\globalmem}{SGEB\xspace}
\newcommand{\Title}{\ours: Question-Guided Geometric Memory for Video Spatial Reasoning}

\title{\Title}
\date{}

\author[1,2]{%
	{\hspace{1mm}Xianqiang Gao\thanks{Equal contribution}}%
}
\author[2]{%
	{\hspace{1mm}Qizhi Chen\protect\footnotemark[1]}%
}
\author[2]{%
	{\hspace{1mm}Delin Qu}%
}
\author[2]{%
	{\hspace{1mm}Haoming Song}%
}
\author[2]{%
	\authorcr
	\vspace{0.3em}
	{\hspace{1mm}Zhigang Wang}%
}
\author[2,3]{%
	{\hspace{1mm}Bin Zhao}%
}
\author[2]{%
	{\hspace{1mm}Dong Wang}%
}
\author[4]{%
	{\hspace{1mm}Xuelong Li}%
}

\affil[1]{University of Science and Technology of China}
\affil[2]{Shanghai AI Lab}
\makeatletter
\renewcommand\AB@affilsep{\protect\\\protect\Affilfont}
\makeatother
\affil[3]{Northwestern Polytechnical University}
\affil[4]{TeleAI}

\renewcommand{\shorttitle}{}

\hypersetup{
  pdftitle={\Title},
  pdfsubject={VLM, Video Spatial Reasoning},
  pdfauthor={Xianqiang Gao, Qizhi Chen, Delin Qu, Haoming Song, Zhigang Wang, Bin Zhao, Dong Wang, Xuelong Li},
  colorlinks=true,
  citecolor=navyblue,
  linkcolor=navyblue,
  urlcolor=navyblue
}

\begin{document}
\maketitle

\vspace{-5em}
\begin{center}
\small \url{https://q-geomem.github.io}
\end{center}
\vspace{1.0em}

\begin{figure}[h]
    \centering
    \includegraphics[width=\linewidth]{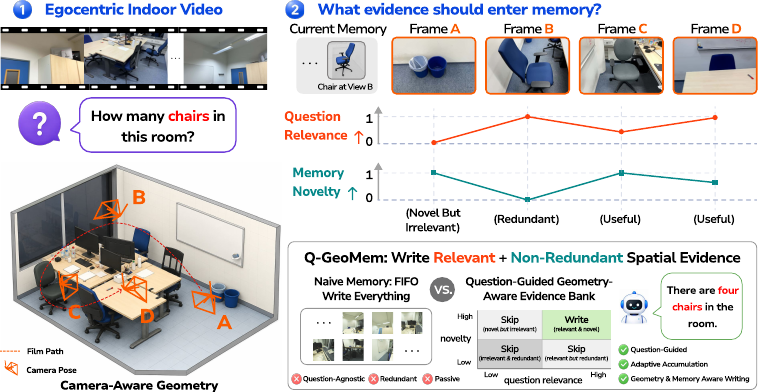}
    \caption{\textbf{Motivation of \ours.} Egocentric indoor videos reveal spatial layout through partial and camera-dependent views, so long-horizon spatial reasoning depends on retaining the right evidence rather than simply storing more frames. For a question such as ``How many chairs are in this room?'', a chronological memory may mix useful chair observations with irrelevant or repeated views. \ours instead treats memory update as question-guided geometric evidence management: camera-conditioned geometry grounds frame features, question relevance identifies task-useful observations, and bank-relative novelty discourages redundant long-range evidence.}
    \label{fig:teaser}
\end{figure}

\vspace{0.5em}

\begin{abstract}

Video spatial reasoning requires accumulating viewpoint-dependent evidence over time while retaining information useful to the question being asked. Existing spatial video-language models improve geometric perception and long-range context modeling, but often treat memory as a generic temporal cache, which can introduce redundant or irrelevant evidence and weaken long-horizon reasoning. We propose \textbf{\ours}, a question-guided geometric memory framework for video spatial reasoning. \ours injects camera-conditioned geometry into visual tokens and maintains two complementary memories: a Fine-Grained Context Bank for recent dense features and camera states, and a Semantic-Geometric Evidence Bank for compact long-range evidence. For each candidate frame, a calibrated Q-Former estimates question relevance, while novelty and evidence utility are recomputed with respect to the active evidence bank. The resulting relevance--novelty utility controls capacity-based replacement and serves as an attention bias during memory reading. During reasoning, both memories are read before update and adaptively fused with the current frame representation. Extensive experiments across two in-domain and five out-of-distribution benchmarks, and controlled memory analyses show that \ours achieves state-of-the-art performance in the evaluated settings and validate the effectiveness of question-guided geometric evidence selection.

\end{abstract}

\section{Introduction}

Memory turns observation into understanding. Perception alone offers only fleeting fragments of the world, while understanding and reasoning emerge from retaining what has been seen, integrating it with new evidence and preserving a coherent representation of the scene. This requirement is especially important in video spatial reasoning, where object relations, camera motion, distances, and directions are revealed gradually from viewpoint-dependent visual evidence.

Recent progress in video spatial reasoning has been driven by stronger benchmarks and increasingly spatially grounded video-language models. Benchmarks such as~\cite{yangThinkingSpaceHow2025,yangMMSIBenchBenchmarkMultiImage2025}, together with spatial-temporal evaluation suites such as~\cite{liSTIBenchAreMLLMs2025,jiaOmniSpatialComprehensiveSpatial2026}, show that video-based spatial intelligence requires more than short-range perception: models must preserve viewpoint-dependent observations, integrate geometric cues over time, and reason about order, distance, and direction from partial visual evidence. In response, recent methods have advanced along three directions: geometry-grounded perception and fusion~\cite{zhengLearningVideos3D2025,fanVLM3RVisionLanguageModels2025,zhaoSpaceMindCameraGuidedModality2025,liThinkingGeometryActive2026a}, spatial reasoning supervision and curricula~\cite{ouyangSpaceRReinforcingMLLMs2025,li2025spatialladder}, and memory-augmented temporal reasoning~\cite{liuVisionLanguageMemorySpatial2025,qian2024streaming,xiong2025streaming,zhangFlashVStreamEfficientRealtime2025,qianDispiderEnablingVideo2025}. Together, these methods improve how spatial evidence is perceived, trained, and retained.

Despite this progress, a central difficulty remains only partially solved: \textbf{\emph{what spatial evidence should be remembered, in what representation, and under what update criterion}}. Geometry-enhanced methods have improved how spatial evidence is formed. VLM-3R~\cite{fanVLM3RVisionLanguageModels2025} aligns monocular videos with instruction-aligned 3D reconstruction, grounding video reasoning in reconstructive 3D priors. SpaceMind~\cite{zhaoSpaceMindCameraGuidedModality2025} shows that camera-guided fusion is more effective than naive multimodal fusion. However, these methods mainly improve evidence formation or fusion, while leaving open how evidence should be retained for a specific question. Memory-oriented methods address temporal retention more directly. VLM$^2$~\cite{liuVisionLanguageMemorySpatial2025} demonstrates that persistent memory is valuable for long-horizon spatial reasoning, yet prior memory-augmented designs often use memory primarily as a repository of generic history rather than explicitly deciding, before writing, whether a candidate observation is relevant to the question and complementary to existing evidence. A similar gap appears in broader long-video methods~\cite{qian2024streaming,xiong2025streaming,zhangFlashVStreamEfficientRealtime2025,qianDispiderEnablingVideo2025}, which maintain history but are not designed around camera-conditioned spatial evidence. As a result, current systems can preserve more temporal context without necessarily preserving the evidence that is most useful for the spatial question being asked.

We argue that this coupling between question relevance, geometric reliability, and memory update is the missing piece. As illustrated in \cref{fig:teaser}, video spatial reasoning requires memory to preserve question-relevant and non-redundant spatial evidence rather than passively cache chronological observations. In video spatial reasoning, memory should therefore function as a question-guided evidence manager: camera-conditioned geometry improves the reliability of each observation, the question determines which observations are useful, and novelty with respect to existing memory discourages storing redundant evidence. The key challenge is not simply to remember more video, but to construct a compact memory that retains sufficient visual-spatial evidence for answering the question.

Motivated by this perspective, we propose \textbf{\ours}, a memory-centric architecture for video spatial reasoning. \ours injects camera-conditioned geometry into visual tokens to obtain spatially grounded frame features, then maintains two complementary memory banks. The Fine-Grained Context Bank (\localmem) keeps recent dense features and camera states for local, view-dependent details, while the Semantic-Geometric Evidence Bank (\globalmem) stores compact long-range evidence with calibrated Q-Former-based question relevance. During reasoning, both banks are read before update and adaptively fused into the current feature. \globalmem recomputes relevance--novelty utility against the active bank for read-time attention bias and capacity-based replacement.

Our contributions are three-fold:
\begin{itemize}
    \item We introduce a \textbf{memory-centric formulation} of video spatial reasoning, in which camera-conditioned geometry, question relevance, and novelty jointly determine what evidence should be retained.
    \item We propose \textbf{\ours}, a unified architecture that combines camera-guided geometry fusion, a Fine-Grained Context Bank for recent dense evidence, a Semantic-Geometric Evidence Bank for compact long-range evidence, and adaptive fusion of memory readouts.
    \item Extensive experiments and ablations show that selecting memory entries by both question relevance and novelty yields a more effective memory mechanism for video spatial reasoning.
\end{itemize}

\section{Related Work}

\textbf{MLLMs for Video Understanding and Reasoning.}
Recent video MLLMs have advanced from short-clip perception toward longer-horizon reasoning by improving how temporally distributed evidence is selected, organized, and queried before answering.
A common trend is to replace uniform frame sampling with question- or task-aware frame selection~\cite{buch2025flexible,hu2025mllm,tangAdaptiveKeyframeSampling2025}, showing that many video questions can be answered from a compact subset of informative frames.
Beyond frame selection, hierarchical querying methods such as HierarQ~\cite{azad2025hierarq} organize temporal evidence at multiple granularities, enabling stronger long-range reasoning without explicitly maintaining an external memory bank.
These methods improve temporal evidence acquisition for general video understanding.
However, they are not designed around viewpoint-dependent spatial evidence: they do not explicitly model how question relevance should interact with spatial geometry, nor how such evidence should be retained once the relevant frame has passed.

\textbf{Geometry-Augmented Video Spatial Reasoning.}
Video spatial reasoning increasingly relies on geometric priors, camera motion, and reconstruction cues.
VG-LLM~\cite{zhengLearningVideos3D2025} injects video-derived 3D geometry priors into MLLMs by fusing geometric cues with visual tokens, showing that implicit geometry can improve video spatial reasoning.
VLM-3R~\cite{fanVLM3RVisionLanguageModels2025} further couples geometry-aware modeling with reconstructive instruction data and the VSTI-Bench evaluation suite, providing both a strong training recipe and a benchmark setting for spatial-temporal reasoning under camera motion.

Recent work has moved beyond simply exposing geometry to the model, and instead studies how geometry should be supervised, fused, or selected.
SpaceR~\cite{ouyangSpaceRReinforcingMLLMs2025} strengthens video spatial reasoning through spatially targeted reinforcement-style training, while SpatialLadder~\cite{li2025spatialladder} studies progressive curricula for improving spatial competence.
SpaceMind~\cite{zhaoSpaceMindCameraGuidedModality2025} shows that camera-guided modality fusion is more effective than shallow multimodal concatenation, emphasizing the role of camera state in controlling spatial evidence formation.
GeoThinker~\cite{liThinkingGeometryActive2026a} further argues for active geometry integration rather than indiscriminate geometry fusion.
Related embodied and robotic studies also explore grounding spatial or affordance-relevant cues from visual observations and using more deliberative reasoning for spatial decision making~\cite{gao2025learning,song2025hume}.
Together, these works improve how spatial evidence is formed, supervised, or fused. Our work addresses a complementary question: once camera-calibrated spatial evidence is formed, which observations should be retained over time for the question being asked?

\textbf{Memory-Augmented Video Understanding.}
For long-horizon video reasoning, memory-based methods differ in what they carry across time. One line maintains explicit historical stores, where selected observations are preserved in an addressable memory and updated online~\cite{xiong2025streaming,huang2025online,zhangFlashVStreamEfficientRealtime2025,dikoReWindUnderstandingLong2025}. Another line represents history as a latent recurrent state rather than an explicit bank, allowing temporally accumulated context to be propagated in a bounded hidden form~\cite{wangVideoLLaMBLongStreaming2025,ren2025vamba,shenSimpleBaselineStreaming2026}. These designs are effective for general long-video understanding, but their memories are usually not specialized for camera-conditioned spatial evidence.

The closest line to our work designs memory specifically for spatial reasoning over video. VLM$^2$~\cite{liuVisionLanguageMemorySpatial2025} introduces a dual-memory architecture with working memory for recent context and episodic memory for long-horizon recall, showing that persistent memory is important for VSI-style spatial reasoning. However, its memory update is primarily driven by learned saliency and similarity-based replacement, and the retained memory is not explicitly selected according to the question being asked. In contrast, \ours treats long-term memory as an active evidence bank: each entry is evaluated by calibrated question relevance and bank-relative novelty, and the resulting utility is used consistently for read-time attention bias and capacity-based replacement. This makes the memory update question-conditioned and geometry-aware, rather than a generic temporal accumulation mechanism.

Overall, prior work highlights the importance of geometry and memory for video spatial reasoning, but often separates spatial evidence formation from question-specific memory update. \ours bridges this gap by retaining compact, camera-calibrated evidence that is both relevant to the question and non-redundant over time.

\section{Method}
\label{sec:method}

\begin{figure}[t]
    \centering
    \includegraphics[width=\linewidth]{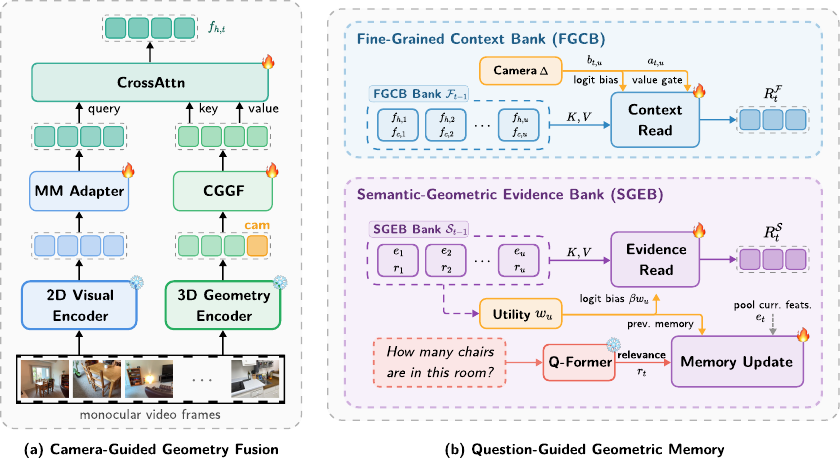}
    \caption{\textbf{Overview of the proposed framework.} Camera-guided geometry fusion first injects spatial cues into frame tokens. The Fine-Grained Context Bank preserves recent detailed visual evidence, while the Semantic-Geometric Evidence Bank stores compact pooled fused features. Question-conditioned utilities are recomputed from the current question and the active evidence bank, enabling memory-aware evidence reading and replacement.}
    \label{fig:method}
\end{figure}

\subsection{Overview}
Given a monocular video and a question, \ours builds a question-guided, geometry-aware memory for video spatial reasoning.
As shown in \cref{fig:method}, we first calibrate each frame's visual tokens with camera and geometry cues, then maintain a question-conditioned memory of past evidence, and finally fuse the current observation with retrieved evidence for answer generation.
The memory is split into a Fine-Grained Context Bank (\localmem) that preserves recent dense features for short-range details, and a Semantic-Geometric Evidence Bank (\globalmem) that stores compact question-relevant evidence for long-range reasoning.

We describe geometry fusion in \cref{sec:cggf}, memory construction in \cref{sec:memory}, and adaptive fusion in \cref{sec:adaptive-fusion}.

\subsection{Camera-Guided Geometry Fusion}
\label{sec:cggf}

Before memory construction, we fuse visual tokens with camera-conditioned geometry, following the camera-guided fusion of SpaceMind~\cite{zhaoSpaceMindCameraGuidedModality2025} so that both \localmem and \globalmem operate on spatially grounded features.
The motivation is that the same geometric observation may support different spatial relations under different camera states.

For frame $t$, let $f_{v,t}\in\mathbb{R}^{N_v\times d}$, $f_{g,t}\in\mathbb{R}^{N_g\times d_g}$, and $f_{c,t}\in\mathbb{R}^{1\times d_g}$ denote the visual, geometry, and camera tokens, where $f_{c,t}$ is broadcast along $N_g$ when needed.
We condition the geometry on the camera token and predict a geometry-only reliability gate:
\begin{equation}
    \mathbf{b}_g = \mathrm{MLP}_b\bigl([W_g^{g} f_{g,t};\,W_c^{g} f_{c,t}]\bigr),\quad
    \mathbf{r}_g = \sigma\bigl(\mathrm{MLP}_r(W_g^{g} f_{g,t})\bigr),
\end{equation}
where $[\,\cdot\,;\,\cdot\,]$ denotes channel-wise concatenation (with the frame-level camera feature broadcast along $N_g$), $\mathbf{b}_g\in\mathbb{R}^{N_g\times d}$, $\mathbf{r}_g\in\mathbb{R}^{N_g\times 1}$, and projection matrices with superscript $g$ are learnable projections for geometry fusion.
The visual stream then attends to the camera-conditioned geometry representation to obtain a geometry residual:
\begin{equation}
    \hat f_{h,t}
    =
    \operatorname{Attn}\!\left(
        W_q^{g} f_{v,t},
        W_k^{g} f_{g,t}+\mathbf{b}_g,
        \bigl(W_v^{g} f_{g,t}+\mathbf{b}_g\bigr)\odot \mathbf{r}_g
    \right),
\end{equation}
where $\hat f_{h,t}\in\mathbb{R}^{N_v\times d}$.
Finally, a camera-conditioned SwiGLU-style channel gate controls how strongly the geometry residual is injected into the visual features:
\begin{align}
    \mathbf{g}_c &= \operatorname{SwiGLU}\!\left(W_{c,v}^{g} f_{c,t}\right),\\
    f_{h,t} &= f_{v,t} + \mathbf{g}_c
    \odot
    W_o^{g} \hat f_{h,t},
\end{align}
where $\mathbf{g}_c\in\mathbb{R}^{1\times d}$ is broadcast over tokens and $f_{h,t}\in\mathbb{R}^{N_v\times d}$.
The geometry-aware feature $f_{h,t}$ is shared by both memory branches in \cref{sec:memory}.

\subsection{\ours}
\label{sec:memory}

Over the geometry-aware features $f_{h,t}$, \ours maintains two complementary memory branches.
The Fine-Grained Context Bank (\localmem) preserves recent dense features for short-range details, while the Semantic-Geometric Evidence Bank (\globalmem) stores compact question-relevant evidence for long-range reasoning.

In \globalmem, a pretrained Q-Former estimates calibrated question relevance for each frame, while novelty is computed against the active evidence bank. The two terms are multiplied into an evidence utility that is recomputed for reading and replacement, so that question-irrelevant or redundant evidence contributes less to subsequent reads. The bank is kept at a fixed capacity by replacing the entry with the lowest relevance--novelty product when full.

\subsubsection{Fine-Grained Context Bank}
The Fine-Grained Context Bank (\localmem) provides short-range dense context by reading recent fused features and camera variation before the current frame is written.
At step $t$, with context window size $\tau$, we maintain a window-based memory:
\begin{equation}
    \mathcal{F}_{t-1}
    =
    [(f_{h,u},f_{c,u})]_{u=\max(1,t-\tau)}^{t-1},
\end{equation}
where $u$ indexes the recent past frames retained in the buffer.
\paragraph{Memory read.}
Viewpoint changes play an important role in tasks such as camera displacement or movement direction. To modulate the readout by how the current viewpoint differs from each historical entry, we use $u$ to index each retained past frame and compute a per-entry camera-feature difference that biases the attention logits and gates the memory values in the memory read:
\begin{equation}
\begin{gathered}
    \Delta f_{c,t,u}
    =
    f_{c,t}-f_{c,u},\\
    b_{t,u}
    =
    \mathrm{MLP}_b^{\mathcal{F}}(\Delta f_{c,t,u}),
    \quad
    a_{t,u}
    =
    \sigma \bigl(\mathrm{MLP}_a^{\mathcal{F}}(\Delta f_{c,t,u})\bigr),
\end{gathered}
\end{equation}
where $b_{t,u},a_{t,u}\in\mathbb{R}$ are scalars associated with history entry $u$.
The \localmem readout is then computed as camera-modulated attention over the concatenated historical features:
\begin{equation}
    R_t^{\mathcal{F}}
    =
    \operatorname{Attn}\!\left(
        W_q^{\mathcal{F}}f_{h,t},\;
        W_k^{\mathcal{F}}[f_{h,u}]_{\mathcal{F}_{t-1}},\;
        (W_v^{\mathcal{F}}[f_{h,u}]_{\mathcal{F}_{t-1}}) \odot [a_{t,u}]_{\mathcal{F}_{t-1}} \;;\;
        [b_{t,u}]_{\mathcal{F}_{t-1}}
    \right),
\end{equation}
where $R_t^{\mathcal{F}}\in\mathbb{R}^{N_v\times d}$, and the bracketed tensors concatenate all \localmem entries in temporal order.
The gate $[a_{t,u}]_{\mathcal{F}_{t-1}}$ rescales the memory values, and the term after the semicolon $[b_{t,u}]_{\mathcal{F}_{t-1}}$ is added to the attention logits as a per-key bias.

\paragraph{Memory write.}
After reading, \localmem appends $(f_{h,t},f_{c,t})$ and discards the oldest entry when the buffer exceeds $\tau$.
This sliding buffer preserves fine-grained geometry-aware visual features that support local spatial relations and view-dependent details.

\subsubsection{Semantic-Geometric Evidence Bank}

\globalmem stores a fixed-size bank of compact semantic-geometric entries with capacity $K$.
At step $t$, let $\mathcal{K}_{t-1}\subseteq\{1,\dots,t-1\}$ denote the indices of past frames currently retained in the bank, with $|\mathcal{K}_{t-1}|\leq K$, so that each $u\in\mathcal{K}_{t-1}$ refers to one stored entry. We maintain
\begin{equation}
    \mathcal{S}_{t-1}
    =
    [(e_u,r_u)]_{u\in\mathcal{K}_{t-1}},
\end{equation}
where $e_u\in\mathbb{R}^{M \times d}$ is the pooled semantic-geometric entry extracted from frame $u$, and $r_u\in\mathbb{R}$ is its calibrated question-relevance score.

\paragraph{Evidence utility.}
We score each candidate frame by multiplying its question relevance with its novelty relative to an active set, which discourages repeatedly storing redundant evidence.
For frame $i$, we compute its question relevance
\begin{equation}
\label{eq:question-relevance}
    r_i
    =
    \operatorname{Calib}\!\left(
    \operatorname{sim}(\mathcal Q(f_{v,i}), \bar q)
    \right),
\end{equation}
where $\bar q\in\mathbb{R}^{d}$ is the pooled question feature from the text branch of $\mathcal{Q}$, and $\operatorname{Calib}(\cdot)$ is the video-level calibration detailed in \cref{sec:video-level-calibration}.
Relevance is computed per frame, whereas novelty is measured against the active entries used by the current operation.
Let $\mathcal{A}$ denote this active index set: $\mathcal{A}=\mathcal{K}_{t-1}$ for memory reading and $\mathcal{A}=\mathcal{C}_t$ for capacity-based replacement.
For each $u\in\mathcal{A}$, we define
\begin{align}
    \nu_u(\mathcal{A})
    &=
    1-\max_{v\in \mathcal{A},\,v\neq u}
    \operatorname{sim}(e_u,e_v),\\
    w_u(\mathcal{A})
    &=
    r_u\,\nu_u(\mathcal{A}),
\end{align}
so that $\nu_u$ is large when $e_u$ differs from the other active entries, and $w_u$ favors evidence that is simultaneously question-relevant and non-redundant.
When unambiguous, we omit the explicit dependence on $\mathcal{A}$ and write $\nu_u$ and $w_u$.

\paragraph{Memory read.}
At step $t$, \globalmem reads from the existing bank $\mathcal{S}_{t-1}$ before any update.
The current fused feature queries the stored entries, while the evidence utility of each entry is injected as a per-key bias on the attention logits:
\begin{equation}
\label{eq:sgeb-read}
    R_t^{\mathcal{S}}
    =
    \operatorname{Attn}\!\left(
        W_q^{\mathcal{S}}f_{h,t},\;
        W_k^{\mathcal{S}}[e_u]_{\mathcal{S}_{t-1}},\;
        W_v^{\mathcal{S}}[e_u]_{\mathcal{S}_{t-1}}\;;\;
        \beta [w_u]_{\mathcal{S}_{t-1}}
    \right),
\end{equation}
where $R_t^{\mathcal{S}}\in\mathbb{R}^{N_v\times d}$ and the bracketed tensors concatenate all \globalmem entries in temporal order. The term after the semicolon adds the per-key bias $\beta\,w_u$ to the attention logits, where $\beta=\operatorname{softplus}(\theta_h)\geq0$ is a learnable scale for each $w_u$. Thus, the evidence utility softly prioritizes high-value entries in the attention logits without directly scaling the memory values.

\paragraph{Memory write.}
After reading, the candidate entry is presented to the bank. While $|\mathcal{K}_{t-1}|<K$, the candidate is appended. Once the bank is full, we form the candidate set $\mathcal{C}_t=\mathcal{K}_{t-1}\cup\{t\}$, recompute the utility $w_u$ of each entry with novelty measured over $\mathcal{C}_t$, and drop
\begin{equation}
    u^\star=\arg\min_{u\in\mathcal{C}_t} w_u.
\end{equation}
If $u^\star=t$, the candidate is discarded. Otherwise, the lowest-scoring stored entry is evicted and the candidate is inserted. The relevance--novelty product $w_u$ therefore drives both capacity-based replacement and subsequent memory reading (\cref{eq:sgeb-read}) without being stored as a static memory attribute.

\subsection{Adaptive Fusion}
\label{sec:adaptive-fusion}

We fuse the memory readouts into the current feature with channel-wise gates conditioned on mean-pooled summaries (denoted by bars, where each visual summary is pooled along the token dimension to $\mathbb{R}^{1\times d}$):
\begin{equation}
\begin{gathered}
    \mathbf{g}_t^{\mathcal{F}}
    =
    \sigma\!\left(\mathrm{MLP}_g^{\mathcal{F}}\!\left([\bar f_{h,t};\bar R_t^{\mathcal{F}}]\right)\right),
    \quad
    \mathbf{g}_t^{\mathcal{S}}
    =
    \sigma\!\left(\mathrm{MLP}_g^{\mathcal{S}}\!\left([\bar f_{h,t};\bar q;\bar R_t^{\mathcal{F}};\bar R_t^{\mathcal{S}}]\right)\right),
    \\
    \tilde f_{h,t}
    =
    f_{h,t}+\mathbf{g}_t^{\mathcal{F}}\odot R_t^{\mathcal{F}}+\mathbf{g}_t^{\mathcal{S}}\odot R_t^{\mathcal{S}}.
\end{gathered}
\end{equation}
where $\bar q$ is the projected question summary, $\mathbf{g}_t^{\mathcal{F}},\mathbf{g}_t^{\mathcal{S}}\in\mathbb{R}^{1\times d}$ are token-broadcast gates, and the fused frame representations are provided to the multimodal LLM.

\section{Experiments}
\label{sec:experiments}

\subsection{Implementation Details}

\paragraph{Training details.}
\ours is built on LLaVA-NeXT-Video-7B~\cite{zhang2024llavanextvideo} and follows the VLM-3R~\cite{fanVLM3RVisionLanguageModels2025} training recipe for video preprocessing, frame sampling, instruction tuning, optimization, and geometry/camera feature extraction.
We train on VLM3R-Data~\cite{fanVLM3RVisionLanguageModels2025} and VICA-322K~\cite{feng2025visuospatial} for two epochs.
For \ours-specific hyperparameters, we set the \localmem context window to $\tau=4$, the \globalmem capacity to $K=8$, and the number of pooled semantic-geometric tokens per frame to $M=7\!\times\!7$.
Full training and spatial-tower configurations are provided in \cref{sec:appendix-training-configuration}.

\paragraph{Benchmarks and metrics.}
We conduct a comprehensive evaluation of \ours across the main video spatial reasoning setting and several out-of-distribution (OOD) transfer settings.
VSI-Bench~\cite{yangThinkingSpaceHow2025} and VSTI-Bench~\cite{fanVLM3RVisionLanguageModels2025} serve as the video spatial reasoning benchmarks, covering static and temporally evolving spatial questions under camera motion.
To assess OOD generalization beyond the main setting, we further evaluate on five complementary benchmarks, including SPBench-MV~\cite{li2025spatialladder} and MMSI-Video~\cite{lin2025mmsivideobench} for video spatial reasoning, SQA3D~\cite{maSQA3DSituatedQuestion2023a} for situated 3D scene question answering, and VideoMME~\cite{fu2025videomme} and EgoSchema~\cite{mangalam2023egoschema} for general long-form video understanding.
We report results on the MV (multi-view) subset of SPBench, as it better aligns with the design of \ours by evaluating reasoning across multiple views rather than single-frame spatial inference.
VSTI-Bench$^\dagger$ denotes the out-of-domain setting in which the evaluated model is not trained on the VLM-3R VSTI training set.

\paragraph{Baseline methods.}
We compare with representative proprietary VLMs, open-source video VLMs, and spatially specialized video-language models.
The proprietary and open-source models test whether general-purpose video understanding is sufficient for spatial reasoning, while specialized models such as VG-LLM~\cite{zhengLearningVideos3D2025}, VST~\cite{yang2025visual}, VLM-3R~\cite{fanVLM3RVisionLanguageModels2025}, and VLM$^2$~\cite{liuVisionLanguageMemorySpatial2025} incorporate geometry, spatial supervision, or memory mechanisms.
This comparison isolates whether question-guided geometric memory provides benefits beyond stronger backbones, geometry-aware representations, or generic memory storage.

\subsection{Results}

\begin{table*}[t] %
    \centering
    \caption{\textbf{Evaluation on VSI-Bench~\cite{yangThinkingSpaceHow2025}.}
    We report average accuracy and eight task-wise scores for proprietary VLMs, open-source video VLMs, and spatially specialized models. \colorbox{linecolor2}{Best} and \colorbox{linecolor1}{second-best} results are highlighted within each model category. \ours achieves the best average score and shows consistent gains over VLM-3R, indicating that question-guided geometric memory improves the retention of task-relevant spatial evidence.}
    \label{tab:vsibench}
    \setlength{\tabcolsep}{3pt}
    \resizebox{1.0\textwidth}{!}{
        \begin{tabular}{lc|cccccccc}
            \toprule
            \multirow{2}{*}[-0.5ex]{Methods}            & \multirow{2}{*}[-0.5ex]{Avg.} & \multicolumn{4}{c}{Numerical Question} & \multicolumn{4}{c}{Multiple-Choice Question}                                                                                                                                                                                           \\
            \cmidrule(lr){3-6}\cmidrule(lr){7-10}
                                                                 &                               & Obj. Cnt.                                       & Abs. Dist.                                            & Obj. Size                    & Room Size                    & Rel. Dist.                   & Rel. Dir.                    & Route Plan                   & Appr. Order                  \\
            \midrule
            \rowcolor{navyblue!5}
            \multicolumn{10}{l}{\textcolor{black}{\textit{Proprietary Models (API)}}}                                                                                                                                                                                                                                                                                                                \\
            GPT-4o~\cite{hurst2024gpt4o}                         & 34.0                          & 46.2                                            & 5.3                                                   & 43.8                         & 38.2                         & 37.0                         & 41.3                         & 31.5                         & 28.5                         \\
            Gemini-1.5 Flash~\cite{team2024gemini}               & 42.1                          & 49.8                                            & 30.8                                                  & 53.5                         & 54.4                         & 37.7                         & 41.0                         & 31.5                         & 37.8                         \\
            Gemini-1.5 Pro~\cite{team2024gemini}                 & 45.4                          & 56.2                                            & 30.9                                                  & 64.1                         & 43.6                         & 51.3                         & 46.3                         & 36.0                         & 34.6                         \\
            \midrule
            \rowcolor{navyblue!5}
            \multicolumn{10}{l}{\textcolor{black}{\textit{Open-sourced VLMs}}}                                                                                                                                                                                                                                                                                                                       \\
            LLaVA-NeXT-Video-7B~\cite{zhang2024llavanextvideo}   & 35.6                          & 48.5                                            & 14.0                                                  & 47.8                         & 24.2                         & \cellcolor{linecolor1}{43.5} & \cellcolor{linecolor2}{42.4} & \cellcolor{linecolor1}{34.0} & 30.6                         \\
            LLaVA-NeXT-Video-72B~\cite{zhang2024llavanextvideo}  & \cellcolor{linecolor1}{40.9}  & \cellcolor{linecolor1}{48.9}                    & 22.8                                                  & \cellcolor{linecolor1}{57.4} & 35.3                         & 42.4                         & 36.7                         & \cellcolor{linecolor2}{35.0} & \cellcolor{linecolor1}{48.6} \\
            Qwen2.5VL-7B~\cite{baiQwen25VLTechnicalReport2025a}  & 33.0                          & 40.9                                            & 14.8                                                  & 43.4                         & 10.7                         & 38.6                         & 38.5                         & 33.0                         & 29.8                         \\
            LLaVA-OneVision-7B~\cite{li2024llavaonevision}       & 32.4                          & 47.7                                            & 20.2                                                  & 47.4                         & 12.3                         & 42.5                         & 35.2                         & 29.4                         & 24.4                         \\
            LLaVA-OneVision-72B~\cite{li2024llavaonevision}      & 40.2                          & 43.5                                            & \cellcolor{linecolor1}{23.9}                          & \cellcolor{linecolor2}{57.6} & \cellcolor{linecolor1}{37.5} & 42.5                         & \cellcolor{linecolor1}{39.9} & 32.5                         & 44.6                         \\
            InternVL3-78B~\cite{zhu2025internvl3}                & \cellcolor{linecolor2}{48.5}  & \cellcolor{linecolor2}{71.2}                    & \cellcolor{linecolor2}{53.7}                          & 44.4                         & \cellcolor{linecolor2}{39.5} & \cellcolor{linecolor2}{55.9} & 39.5                         & 28.9                         & \cellcolor{linecolor2}{54.5} \\
            \midrule
            \rowcolor{navyblue!5}
            \multicolumn{10}{l}{\textcolor{black}{\textit{Specialized Spatial Reasoning Models}}}                                                                                                                                                                                                                                                                                                    \\
            VG-LLM-8B~\cite{zhengLearningVideos3D2025}           & 50.7                          & 67.9                                            & 37.7                                                  & 58.6                         & 62.0                         & 46.6                         & 40.7                         & 32.4                         & 59.2                         \\
            VST-7B~\cite{yang2025visual}                         & 60.6                          & 72.0                                            & 44.4                                                  & \cellcolor{linecolor1}{74.3} & 68.3                         & 59.7                         & 55.8                         & 44.9                         & 65.2                         \\
            VLM-3R-7B~\cite{fanVLM3RVisionLanguageModels2025}    & 60.9                          & 70.2                                            & 49.4                                                  & 69.2                         & 67.1                         & 65.4                         & 80.5                         & 45.4                         & 40.1                         \\
            VLM$^2$-7B~\cite{liuVisionLanguageMemorySpatial2025} & \cellcolor{linecolor1}{68.8}  & \cellcolor{linecolor1}{72.5}                    & \cellcolor{linecolor1}{59.6}                          & 70.8                         & \cellcolor{linecolor1}{69.9} & \cellcolor{linecolor1}{69.0} & \cellcolor{linecolor2}{87.8} & \cellcolor{linecolor2}{52.6} & \cellcolor{linecolor1}{68.3} \\
            \textbf{\ours (Ours)}        & \cellcolor{linecolor2}{71.2}  & \cellcolor{linecolor2}{74.8}                    & \cellcolor{linecolor2}{59.9}                          & \cellcolor{linecolor2}{76.5} & \cellcolor{linecolor2}{75.3} & \cellcolor{linecolor2}{71.1} & \cellcolor{linecolor1}{87.2} & \cellcolor{linecolor1}{50.0} & \cellcolor{linecolor2}{74.6} \\
            \midrule
            {\color{gain}\textit{Improvement over VLM-3R}}       & {\color{gain}+10.3}           & {\color{gain}+4.6}                              & {\color{gain}+10.5}                                   & {\color{gain}+7.3}           & {\color{gain}+8.2}           & {\color{gain}+5.7}           & {\color{gain}+6.7}           & {\color{gain}+4.6}           & {\color{gain}+34.5}          \\
            \bottomrule
        \end{tabular}
    } %
\vspace{-1em}
\end{table*}

\paragraph{VSI-Bench.}
\cref{tab:vsibench} reports the main comparison on VSI-Bench, which covers both numerical spatial estimation and multiple-choice spatial reasoning.
\ours achieves an average score of 71.2, outperforming the strongest spatial-memory baseline VLM$^2$-7B by 2.4 points and exceeding VLM-3R-7B by 10.3 points.
Compared with VLM-3R, \ours improves all eight task categories, with particularly large gains on appearance order (+34.5), absolute distance (+10.5), room-size estimation (+8.2), and object-size estimation (+7.3).
These tasks require aggregating observations from different viewpoints and retaining evidence that may be separated by long temporal intervals, making them sensitive to how memory is written and read.

Compared with VLM$^2$-7B, \ours improves six of the eight tasks, including object counting, absolute distance, object-size estimation, room-size estimation, relative distance, and appearance order, while remaining slightly lower on relative direction and route planning.
The gains are concentrated on questions that require complementary evidence across time, whereas route-level planning remains more difficult and may require more explicit path reasoning.
This comparison reinforces the main motivation of \ours, as effective memory for video spatial reasoning should select question-relevant and geometrically complementary evidence rather than simply expand temporal context.

\begin{table*}[t]
    \centering
    \caption{\textbf{Evaluation on VSTI-Bench~\cite{fanVLM3RVisionLanguageModels2025}.}
    We compare representative proprietary models, open-source video VLMs, and spatially specialized models on five temporal spatial reasoning tasks. \colorbox{linecolor2}{Best} and \colorbox{linecolor1}{second-best} results are highlighted within each model category. \ours achieves the best performance, showing the benefit of coupling camera-aware memory readout with question-guided evidence writing.}
    \label{tab:vstibench}
    \setlength{\tabcolsep}{3pt}
    \resizebox{1.0\textwidth}{!}{
        \begin{tabular}{lc|ccccc}
            \toprule
            \multirow{2}{*}[-0.5ex]{Methods}            & \multirow{2}{*}[-0.5ex]{Avg.} & \multicolumn{2}{c}{Numerical Question} & \multicolumn{3}{c}{Multiple-Choice Question}                                                                                              \\
            \cmidrule(lr){3-4}\cmidrule(lr){5-7}
                                                                 &                                        & Cam-Obj Abs. Dist.                              & Cam. Displace.                                        & Cam. Mov. Dir.               & Obj-Obj Rel. Pos.            & Cam-Obj Rel. Dist.           \\
            \midrule
            \rowcolor{navyblue!5}
            \multicolumn{7}{l}{\textcolor{black}{\textit{Proprietary Models (API)}}}                                                                                                                                                                                                                             \\
            GPT-4o~\cite{hurst2024gpt4o}                         & 38.2                                   & 29.5                                            & 23.4                                                  & 37.3                         & 58.1                         & 42.5                         \\
            Gemini-1.5 Flash~\cite{team2024gemini}               & 32.1                                   & 28.5                                            & 20.9                                                  & 24.4                         & 52.6                         & 33.9                         \\

            \midrule
            \rowcolor{navyblue!5}
            \multicolumn{7}{l}{\textcolor{black}{\textit{Open-sourced VLMs}}}                                                                                                                                                                                                                                    \\
            LLaVA-NeXT-Video-7B~\cite{zhang2024llavanextvideo}   & 40.0                                   & 28.2                                            & 1.8                                                   & \cellcolor{linecolor2}{49.8} & 64.7                         & \cellcolor{linecolor2}{55.6} \\
            LLaVA-OneVision-7B~\cite{li2024llavaonevision}       & 41.7                                   & 29.9                                            & \cellcolor{linecolor1}{19.3}                          & 47.5                         & 62.1                         & 49.8                         \\
            LongVA-7B~\cite{zhang2024long}                       & 32.3                                   & 13.5                                            & 5.1                                                   & 43.7                         & 57.9                         & 41.2                         \\
            InternVL2-8B~\cite{chen2024how}                      & \cellcolor{linecolor1}{43.5}           & \cellcolor{linecolor2}{32.9}                    & 13.5                                                  & 48.0                         & \cellcolor{linecolor1}{68.0} & \cellcolor{linecolor1}{55.0} \\
            LongVILA-8B~\cite{chen2024longvila}                  & 30.5                                   & 20.0                                            & 11.6                                                  & 35.4                         & 52.3                         & 33.4                         \\
            VILA-1.5-8B~\cite{lin2024vila}                       & 37.3                                   & 30.1                                            & \cellcolor{linecolor2}{27.3}                          & 42.2                         & 50.4                         & 36.7                         \\
            VILA-1.5-40B~\cite{lin2024vila}                      & 38.2                                   & 28.2                                            & 15.7                                                  & 28.8                         & 65.4                         & 53.0                         \\
            LLaVA-NeXT-Video-72B~\cite{zhang2024llavanextvideo}  & \cellcolor{linecolor2}{44.0}           & \cellcolor{linecolor1}{32.3}                    & 10.5                                                  & \cellcolor{linecolor1}{48.1} & \cellcolor{linecolor2}{78.3} & 50.9                         \\
            \midrule
            \rowcolor{navyblue!5}
            \multicolumn{7}{l}{\textcolor{black}{\textit{Specialized Spatial Reasoning Models}}}                                                                                                                                                                                                                 \\
            VLM-3R-7B~\cite{fanVLM3RVisionLanguageModels2025}    & 58.8                                   & 39.4                                            & 39.6                                                  & 60.6                         & 86.5                         & 68.6                         \\
            VLM$^2$-7B~\cite{liuVisionLanguageMemorySpatial2025} & \cellcolor{linecolor1}{65.3}           & \cellcolor{linecolor1}{43.1}                    & \cellcolor{linecolor1}{44.1}                          & \cellcolor{linecolor1}{76.8} & \cellcolor{linecolor1}{87.7} & \cellcolor{linecolor2}{74.9} \\
            \textbf{\ours (Ours)}                                & \cellcolor{linecolor2}{67.8}           & \cellcolor{linecolor2}{44.3}                    & \cellcolor{linecolor2}{44.6}                          & \cellcolor{linecolor2}{84.0} & \cellcolor{linecolor2}{91.0} & \cellcolor{linecolor1}{74.9} \\
            \bottomrule
        \end{tabular}
    }
\vspace{-1em}
\end{table*}

\paragraph{VSTI-Bench.}
For this setting, we additionally fine-tune \ours on the VSTI training data and evaluate temporal spatial reasoning under changing camera viewpoints.
As shown in \cref{tab:vstibench}, \ours obtains an average score of 67.8, improving over VLM$^2$-7B by 2.5 points and VLM-3R-7B by 9.0 points, with the best results on four of the five task categories.
The largest gain over VLM$^2$-7B appears on camera movement direction (+7.2), a task that directly depends on comparing local viewpoint changes across adjacent observations.
This improvement aligns with the \localmem read-modulation ablation in \cref{tab:memory_local_ablation}, where Camera-$\Delta$ improves \localmem readout by injecting camera variation into local memory retrieval.
The VSTI-Bench comparison therefore points to complementary roles for the two memories, with Camera-$\Delta$-aware local memory capturing short-range viewpoint changes and question-guided evidence writing supporting longer-range retention.

\begin{table*}[t]
    \centering
    \caption{\textbf{Out-of-distribution generalization.}
    We evaluate the effect of \globalmem on five OOD benchmarks spanning video spatial reasoning, situated 3D scene QA, and general video understanding. All results use the official metric of each benchmark.}
    \label{tab:ood_generalization}
    \small
    \setlength{\tabcolsep}{5pt}

    \begin{tabular}{lccccc}
    \toprule
    Model & SPBench-MV & MMSI-Video & SQA3D & VideoMME & EgoSchema \\
    \midrule
    \ours (w/o \globalmem) & 69.03 & 23.97 & 49.43 & 59.77 & 49.81 \\
    \rowcolor{linecolor2}
    \ours & 72.61 & 25.49 & 50.38 & 60.96 & 53.11 \\
    \bottomrule
    \end{tabular}
\end{table*}

\paragraph{Out-of-distribution generalization.}
To test whether the proposed memory design transfers beyond the main settings, we report OOD results in \cref{tab:ood_generalization}.
All five benchmarks in this evaluation are OOD. SPBench-MV and MMSI-Video test video spatial reasoning, SQA3D tests transfer to situated 3D scene question answering, and VideoMME and EgoSchema test transfer to general video understanding.
Across all five OOD benchmarks, adding \globalmem consistently improves over the variant without \globalmem, with gains of +3.58 on SPBench-MV, +1.52 on MMSI-Video, +0.95 on SQA3D, +1.19 on VideoMME, and +3.30 on EgoSchema.
The largest gains appear on SPBench-MV and EgoSchema, covering both spatially focused transfer and long-form egocentric video QA, where relevant evidence may be sparsely distributed across the sequence.
This transfer behavior is consistent with the role of question-guided geometric memory as a mechanism for retaining task-relevant evidence across time.

\subsection{Ablation studies}

\begin{table*}[ht]
\caption{\textbf{Core component ablation on VSI-Bench.}
Starting from the fine-tuned LLaVA-NeXT-Video-7B baseline, we progressively add camera-guided geometry fusion (CGGF), \localmem, and \globalmem. Results include average accuracy and task-wise scores for numerical and multiple-choice spatial reasoning.}
\label{tab:core_ablation}
\centering
\small
\setlength{\tabcolsep}{4pt}
\resizebox{1.0\textwidth}{!}{
\begin{tabular}{l c cccc cccc}
\toprule
\multirow{2}{*}[-0.5ex]{Model} & \multirow{2}{*}[-0.5ex]{Avg.} & \multicolumn{4}{c}{Numerical Question} & \multicolumn{4}{c}{Multiple-Choice Question} \\
\cmidrule(lr){3-6}\cmidrule(lr){7-10}
& & Obj. Cnt. & Abs. Dist. & Obj. Size & Room Size & Rel. Dist. & Rel. Dir. & Route Plan & Appr. Order \\
\midrule
LLaVA-NeXT-Video-7B ft. (Baseline) & 64.83 & 71.91 & 51.47 & 75.38 & 70.56 & 67.32 & 72.09 & 42.27 & 67.64 \\
Baseline + CGGF & 67.28 & 72.07 & 56.13 & 74.49 & 72.43 & 68.17 & 81.47 & 45.36 & 68.12 \\
Baseline + CGGF + \localmem & 67.92 & 71.70 & 55.97 & 75.47 & 73.75 & 66.48 & 84.62 & 46.91 & 68.45 \\
\rowcolor{linecolor2}
Baseline + CGGF + \localmem + \globalmem & 71.17 & 74.76 & 59.89 & 76.55 & 75.28 & 71.13 & 87.18 & 50.00 & 74.60 \\
\bottomrule
\end{tabular}
}
\vspace{-1em}
\end{table*}

\paragraph{Core components.}
We first ablate the core components to verify how each design choice contributes to video spatial reasoning.
The VSI-Bench results are summarized in \cref{tab:core_ablation}.
Adding camera-guided geometry fusion improves the baseline from 64.83 to 67.28, confirming that geometry is more useful when calibrated by camera state before memory construction.
Introducing \localmem further raises the average to 67.92, mainly by improving tasks that benefit from recent fine-grained visual context, such as relative direction.
The full model with both \localmem and \globalmem reaches 71.17, giving a 6.34-point improvement over the baseline.
Compared with the +CGGF+\localmem variant, adding \globalmem yields especially large gains on appearance order, where the answer depends on retaining complementary evidence across a longer temporal horizon.

\begin{table*}[ht]
    \caption{\textbf{Ablation of \localmem read modulation.}
    We compare \localmem with and without Camera-$\Delta$ while disabling \globalmem, and report results by video length on VSI-Bench and VSTI-Bench$^\dagger$.}
    \label{tab:memory_local_ablation}
    \centering
    \small
    \setlength{\tabcolsep}{12pt}
    \resizebox{\textwidth}{!}{%
    \begin{tabular}{l c c c c c c c c}
    \toprule
    \multirow{2}{*}{\textbf{Model}} & \multicolumn{4}{c}{VSI-Bench} & \multicolumn{4}{c}{VSTI-Bench$^\dagger$} \\
    \cmidrule(lr){2-5}\cmidrule(lr){6-9}
     & Avg. & Short & Mid & Long & Avg. & Short & Mid & Long \\
    \midrule
    \ours (w/o \globalmem) w/o Camera-$\Delta$ & 66.57 & 68.91 & 67.66 & 64.50 & 51.17 & 51.08 & 51.39 & 51.60 \\
    \rowcolor{linecolor2}
     \ours (w/o \globalmem) & 67.92 & 70.55 & 69.43 & 65.33 & 52.24 & 53.25 & 52.16 & 52.32 \\
    \midrule
    {\color{gain}\textit{Improvement}} & {\color{gain}+1.35} & {\color{gain}+1.64} & {\color{gain}+1.77} & {\color{gain}+0.83} & {\color{gain}+1.07} & {\color{gain}+2.17} & {\color{gain}+0.77} & {\color{gain}+0.72} \\
    \bottomrule
    \end{tabular}
    }
    \vspace{-1em}
    \end{table*}

\paragraph{Effect of camera modulation.}
To assess whether camera-state variation provides a useful conditioning signal for \localmem readout, we compare models with and without Camera-$\Delta$ in \cref{tab:memory_local_ablation}.
To isolate this factor, \globalmem is disabled in both variants, and the ablated model removes Camera-$\Delta$ by fixing the per-entry bias and gate to $b_{t,u}=0$ and $a_{t,u}=1$.
This reduces \localmem reading to standard cross-attention over the recent buffer while keeping the remaining architecture unchanged.
Under this controlled comparison, Camera-$\Delta$ improves VSI-Bench from 66.57 to 67.92 and VSTI-Bench$^\dagger$ from 51.17 to 52.24, with positive gains across all length buckets on both benchmarks.
Local camera variation therefore helps the model weight recent observations according to viewpoint changes, providing a complementary signal to the long-range evidence selection analyzed later.

\subsection{Analysis}

\paragraph{Length-based diagnostics.}

\begin{figure}[ht]
    \centering
    \includegraphics[width=\linewidth]{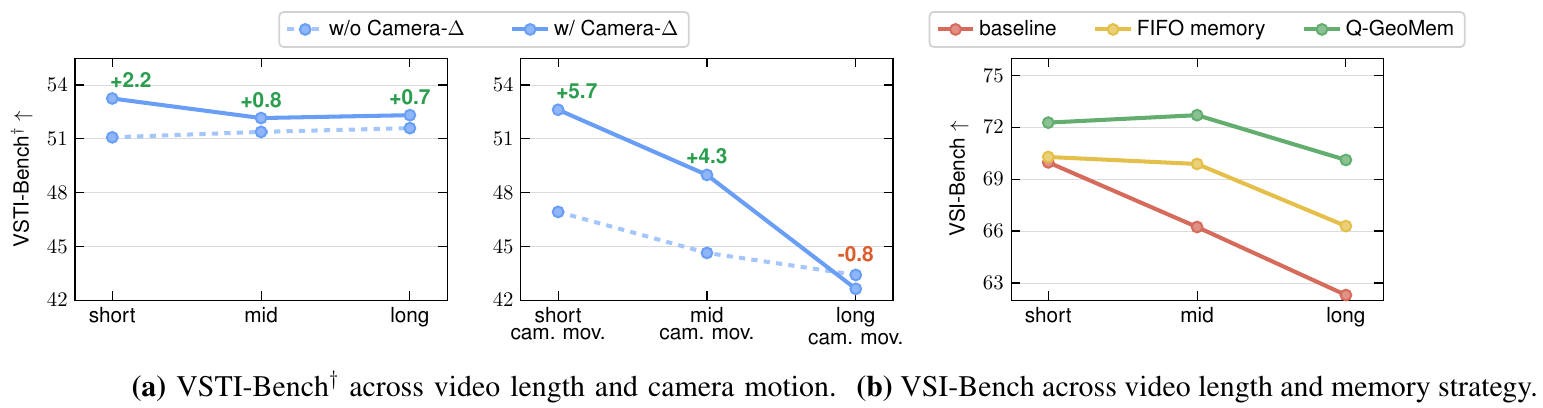}
    \caption{\textbf{Length-based memory diagnostics.} (a) Camera-$\Delta$ modulation improves \localmem readout on VSTI-Bench$^\dagger$, especially for short videos and camera movement direction, while its effect is limited on long videos. (b) The proposed \globalmem design outperforms FIFO memory on VSI-Bench, with larger gains as the video length increases.}
    \label{fig:video_length_diagnostics}
\end{figure}

For a finer-grained diagnosis, \cref{fig:video_length_diagnostics} examines when local camera modulation and long-range evidence selection are most useful.
Panel (a) shows that Camera-$\Delta$ improves short, mid, and long videos on VSTI-Bench$^\dagger$, with the largest gain on short videos.
For the camera movement direction subset, the gains are especially clear on short and mid videos, while the effect is not evident on long videos.
This behavior matches the intended division of labor between the two memories. Camera-$\Delta$ mainly targets local viewpoint transitions, whereas longer videos require an additional mechanism to select sparse but complementary evidence beyond the recent context window.

Panel (b) further compares memory strategies across video length on VSI-Bench.
Compared with FIFO memory, \ours yields larger gains as video length increases. Long-horizon spatial reasoning benefits from actively selecting question-relevant and non-redundant evidence instead of retaining frames only by chronological order.

\paragraph{\globalmem policy and capacity.}
For \globalmem, we examine how active utilities are used for replacement and reading in \cref{tab:memory_ablation}(a), and study how large the evidence bank should be in \cref{tab:memory_ablation}(b).
Under the fixed-capacity setting in \cref{tab:memory_ablation}(a), FIFO reaches 68.7 average accuracy and 66.3 on long videos, while QRel-only and novelty-only scoring provide moderate improvements.
The full relevance--novelty utility achieves 71.2 average accuracy and 70.1 on long videos, improving over FIFO by +2.5 and +3.8 points and over the best single-term policy by +1.9 points on both metrics.
The product utility addresses complementary failure modes. Relevance alone may retain visually similar evidence about the queried content, while novelty alone may preserve diverse but question-irrelevant views.
Combining the two terms favors observations that are both useful to the question and complementary to the current bank.

\cref{tab:memory_ablation}(b) reveals a non-monotonic capacity-accuracy trade-off. The best result is achieved at \(K=8\), which keeps only 25\% of the 32 sampled frames while reaching 71.2 on VSI-Bench and 70.1 on long videos. A smaller bank (\(K=4\)) underperforms, indicating insufficient coverage of complementary evidence, whereas larger banks also fall below \(K=8\), with the full-budget setting \(K=32\) dropping by 3.3 points overall and 3.9 points on long videos. \globalmem works best as a compact evidence bottleneck that retains enough diverse observations while still imposing pressure against redundant or low-utility entries. The larger gain on the long-video split further reflects the value of compact evidence selection in long-horizon spatial reasoning.

\begin{table*}[ht]
\caption{\textbf{Policy analysis of \globalmem on VSI-Bench.}
We vary the \globalmem capacity $K$ and compare FIFO replacement, question-relevance-only (QRel-only) scoring, novelty-only scoring, and the full recomputed relevance--novelty utility used by \ours.}
\label{tab:memory_ablation}
\centering
\small
\begin{minipage}[t]{0.54\textwidth}
\vspace{0pt}
\centering
\textit{(a) \globalmem Memory Policy}\par
\vspace{0.35em}
\setlength{\tabcolsep}{1.5pt}
\begin{tabular}{l l c c c}
\toprule
Policy & Update rule & Read bias & VSI Avg. $\uparrow$ & VSI Long $\uparrow$ \\
\midrule
FIFO & evict oldest & none & 68.7 & 66.3 \\
\rowcolor{linecolor1}
QRel-only & evict $\min r$ & $\beta r$ & 69.3 & 68.2 \\
Novelty-only & evict $\min \nu$ & $\beta \nu$ & 68.9 & 67.3 \\
\rowcolor{linecolor2}
\ours (Ours) & evict $\min r\nu$ & $\beta r\nu$ & 71.2 & 70.1 \\
\bottomrule
\end{tabular}
\end{minipage}%
\hspace{0.04\textwidth}%
\begin{minipage}[t]{0.38\textwidth}
\vspace{0pt}
\centering
\textit{(b) \globalmem Capacity--Accuracy Trade-off}\par
\vspace{0.35em}
\setlength{\tabcolsep}{7pt}
\begin{tabular}{ccc}
\toprule
K & VSI Avg. $\uparrow$ & VSI Long $\uparrow$ \\
\midrule
4 & 69.0 & 67.1 \\
\rowcolor{linecolor2}
8 & 71.2 & 70.1 \\
16 & 68.4 & 66.9 \\
\rowcolor{linecolor1}
24 & 69.5 & 68.3 \\
32 & 67.9 & 66.2 \\
\bottomrule
\end{tabular}
\end{minipage}

\end{table*}

\paragraph{Qualitative evidence selection.}
\begin{figure}[ht]
    \centering
    \includegraphics[width=\linewidth]{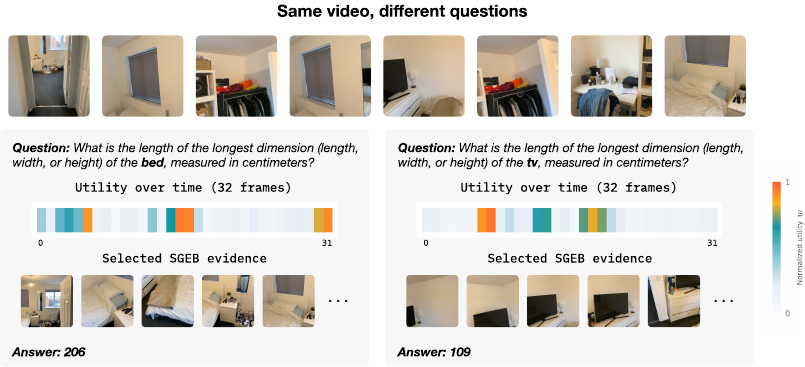}
    \caption{\textbf{Question-conditioned \globalmem evidence selection.}
    The heatmaps show normalized active \globalmem utility \(w=r\nu\) over 32 frames for two questions from the same video, and the selected entries indicate the evidence retained by the bank.}
    \label{fig:mem_illustration}
\end{figure}

To illustrate how \globalmem adapts evidence selection to the question, we visualize qualitative examples in \cref{fig:mem_illustration}. Although the visual stream is identical, changing the queried object from the bed to the TV shifts both the high-utility frames and the retained entries. For the bed question, \ours preserves complementary views of the bed and nearby context, whereas for the TV question, the retained evidence shifts toward frames where the TV is visible. The retained bank is therefore not a fixed chronological cache, but a question-conditioned evidence store that prioritizes relevant and non-redundant observations for the current spatial query.

\section{Conclusion}

We presented \ours, a question-guided geometric memory framework for video spatial reasoning. Instead of using memory as a generic temporal cache, \ours combines camera-conditioned frame calibration with two complementary memories: \localmem for recent dense evidence and \globalmem for compact long-range evidence selected by question relevance and novelty. Experiments on VSI-Bench and VSTI-Bench, together with ablations, show consistent gains over strong geometry-aware and memory-based baselines, especially when reasoning requires temporally distributed observations or camera-motion cues. The findings emphasize that effective video spatial reasoning depends not only on stronger visual representations, but also on explicit selection of the spatial evidence retained for each question.

\paragraph{Limitations.}
Despite these results, \ours has several limitations. The framework relies on extracted camera and geometry cues, so inaccurate pose, depth, or reconstruction signals may weaken memory selection. In addition, our memory mechanism is designed to retain useful spatial evidence, but it does not explicitly model complex multi-step reasoning, such as route-level planning or long chains of spatial transformations. Extending the method to fully online or real-time embodied settings remains an important direction for future work.

\bibliographystyle{plainnat}
\bibliography{ref,ref_auto}

\newpage
\appendix
\section{Implementation Details}
\label{sec:appendix-implementation-details}

\subsection{Training Configuration}
\label{sec:appendix-training-configuration}

\ours is trained with LoRA fine-tuning using rank 128 and alpha 256.
We use per-device batch size 1, gradient accumulation steps 8, learning rate $2\times10^{-5}$, weight decay 0, warmup ratio 0.03, and a cosine learning-rate schedule.
Training uses BF16 and TF32 precision, DeepSpeed ZeRO stage 2, and gradient checkpointing.
Following VLM-3R~\cite{fanVLM3RVisionLanguageModels2025}, we use CUT3R~\cite{wang2025continuous} as the spatial tower, select all spatial features with feature dimension 768, keep the spatial tower frozen, and tune the fusion block and multimodal MLP adapter.
Training takes approximately 38 hours on 16 H200 GPUs.

\subsection{Video-Level Calibration}
\label{sec:video-level-calibration}

The video-level calibration $\operatorname{Calib}(\cdot)$ used in the question-relevance score of \cref{eq:question-relevance} standardizes the raw question--frame similarity within each video, so that relevance is comparable across videos whose overall similarity scales differ.
For frame $i$, let $\rho_i=\operatorname{sim}(\mathcal{Q}(f_{v,i}),\bar q)$ denote the raw similarity between the Q-Former frame embedding and the pooled question feature. We calibrate it as
\[
    r_i
    =
    \sigma\!\left(
    \frac{\rho_i-\mu_\rho}{\sigma_\rho+\epsilon}
    \right),
\]
where $\mu_\rho$ and $\sigma_\rho$ are the mean and standard deviation of $\{\rho_i\}$ computed over the sampled frames of the video, $\epsilon>0$ is a small constant for numerical stability, and $\sigma(\cdot)$ is the logistic sigmoid that maps the standardized score to $(0,1)$.
This per-video standardization removes video-specific similarity offsets and scales, which stabilizes the relevance--novelty utility used for \globalmem reading and replacement.

\subsection{Active Evidence Utility}
\label{sec:active-evidence-utility-details}

As described in \cref{sec:memory}, \globalmem recomputes the relevance--novelty utility over the active entries used by each operation, rather than storing it as a fixed scalar.
For read-time retrieval, the active set is the current bank before the candidate frame is written; for replacement, it is the full candidate set after adding the current frame.
This distinction ensures that retrieval reflects the evidence currently available, while eviction accounts for the redundancy introduced by the new candidate.
When the active set contains only one entry, we set its novelty to 1, so the first retained entry is scored solely by question relevance.
The resulting utility is then used consistently as the read-time attention bias in \cref{eq:sgeb-read} and as the replacement score under the fixed memory budget.

\section{Additional Results}
\label{sec:additional-results}

\subsection{Evidence Utility Usage}
\label{sec:appendix-evidence-utility-usage}

\cref{tab:sgeb_utility_role} isolates where the full relevance--novelty utility \(w=r\nu\) should be used in \globalmem.
The retrieval-off variants remove the utility bias in read attention, while the replacement-off variants retain chronological FIFO eviction.
This 2-by-2 design reports VSI-Bench average accuracy and tests whether the same active utility should jointly control evidence retrieval and memory replacement.

\begin{table}[ht]
\caption{\textbf{Disentangling evidence utility usage.} Starting from the FIFO memory baseline, we fix the full relevance--novelty utility \(w=r\nu\) and enable it either for read-time attention bias, for capacity-based replacement, or for both. We report average accuracy on VSI-Bench, isolating whether the gain comes from better retrieval, better memory composition, or their combination.}
\label{tab:sgeb_utility_role}
\centering
\small
\setlength{\tabcolsep}{3pt}
\begin{tabular}{l c c c}
\toprule
Variants & Active $w$ for replacement & Read bias $\beta w$ & VSI Avg. \\
\midrule
FIFO & \ding{55} & \ding{55} & 68.7 \\
Read bias only & \ding{55} & \checkmark & 69.2 \\
Write replacement only & \checkmark & \ding{55} & 70.2 \\
\rowcolor{linecolor2}
\ours (Ours) & \checkmark & \checkmark & 71.2 \\
\bottomrule
\end{tabular}
\vspace{-1em}
\end{table}

\subsection{Additional Qualitative Results}

\begin{figure}[ht]
    \centering
    \includegraphics[width=\linewidth]{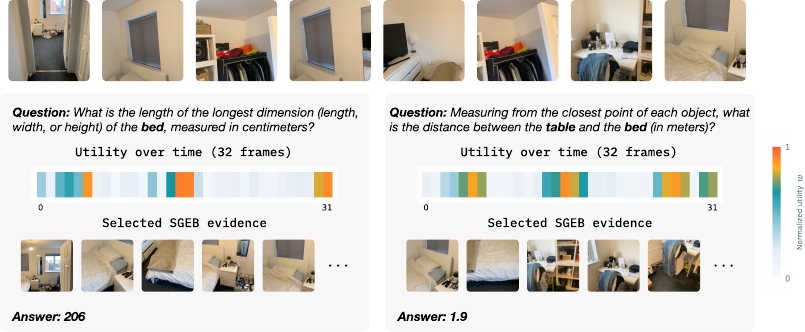}
    \caption{\textbf{Additional question-conditioned \globalmem evidence selection from the same video.}}
    \label{fig:appendix_mem_illustration}
\end{figure}

We provide additional qualitative examples in \cref{fig:appendix_mem_illustration} to further visualize how \globalmem selects question-conditioned evidence from the same video.

\end{document}